\documentclass[letterpaper, 10 pt, conference]{ieeeconf}  % Comment this line out if you need a4paper

\IEEEoverridecommandlockouts                              % This command is only needed if you want to use the \thanks command

\usepackage{amsmath, amssymb, amsfonts}
\usepackage[font=small]{caption}
\usepackage{subcaption}
\usepackage{graphicx}
\usepackage{algorithm2e}
\usepackage{wrapfig}
\usepackage{lipsum}
\usepackage{multirow}
\usepackage{multicol}
\usepackage{cite}
\usepackage{dblfloatfix}
\usepackage{float}
\usepackage{todonotes}
\usepackage{hyperref}

\SetKwInput{KwInput}{Input}
\SetKwInput{KwInit}{Initialize}
\SetKwInput{KwResult}{Output}

\newtheorem{definition}{Definition}
\newtheorem{theorem}{Theorem}

\newtheorem{assumption}{Assumption}

\DeclareMathOperator*{\argmin}{arg\,min}

\newcommand\numberthis{\addtocounter{equation}{1}\tag{\theequation}}

\newcommand{\controllable}[1]{\textcolor{blue}{\textit{#1}}}
\newcommand{\uncontrollable}[1]{\textcolor{red}{\textit{#1}}}

\title{\LARGE \bf
Reactive Temporal Logic-based Planning and Control \\ for Interactive Robotic Tasks
}

\author{Farhad Nawaz$^{1*}$, Shaoting Peng$^{1}$, Lars Lindemann$^{2}$, Nadia Figueroa$^{1}$ and Nikolai Matni$^{1}$% <-this % stops a space
\thanks{$^{1}$Farhad Nawaz, Shaoting Peng, Nadia Figueroa and Nikolai Matni are with the GRASP Lab, University of Pennsylvania, PA 19104, USA.
$^{2}$Lars Lindemann is with the University of Southern California, CA 90007, USA.}
\thanks{$^{*}$Corresponding author: {\tt farhadn@seas.upenn.edu}}
}

\begin{document}

\maketitle
\thispagestyle{empty}
\pagestyle{empty}

\begin{figure*}[!h]
         \centering         \includegraphics[width=\textwidth]{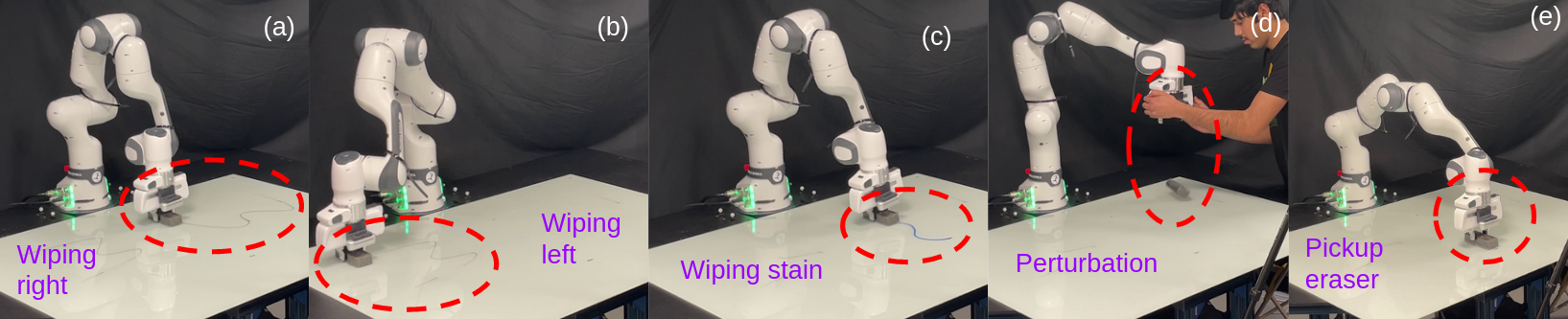}
      \caption{Periodic wiping motion on the (a) right side and (b) left side of the board, (c) wiping the blue stain detected by a camera, (d)~eraser dropped because of human perturbation and (e) robot picks up the dropped eraser.} 
        \vspace{-20pt}
        \label{fig:board-snaps}
\end{figure*}
%%%%%%%%%%%%%%%%%%%%%%%%%%%%%%%%%%%%%%%%%%%%%%%%%%%%%%%%%%%%%%%%%%%%%%%%%%%%%%%%
\begin{abstract}
Robots interacting with humans must be safe, reactive and adapt online to unforeseen environmental and task changes. Achieving these requirements concurrently is a challenge as interactive planners lack formal safety guarantees, while safe motion planners lack flexibility to adapt. To tackle this, we propose a modular control architecture that generates both safe and reactive motion plans for human-robot interaction by integrating temporal logic-based discrete task level plans with continuous Dynamical System (DS)-based motion plans. We formulate a reactive temporal logic formula that enables users to define task specifications through structured language, and propose a planning algorithm at the task level that generates a sequence of desired robot behaviors while being adaptive to environmental changes. At the motion level, we incorporate control Lyapunov functions and control barrier functions to compute stable and safe continuous motion plans for two types of robot behaviors: (i) complex, possibly periodic motions given by autonomous DS and (ii) time-critical tasks specified by Signal Temporal Logic~(STL). Our methodology is demonstrated on the Franka robot arm performing wiping tasks on a whiteboard and a mannequin that is compliant to human interactions and adaptive to environmental changes.
\end{abstract}

%%%%%%%%%%%%%%%%%%%%%%%%%%%%%%%%%%%%%%%%%%%%%%%%%%%%%%%%%%%%%%%%%%%%%%%%%%%%%%%%
\section{INTRODUCTION}
\label{sec:intro}
There is a growing presence of robot assistants in households, workplaces, and industrial plants. In such applications, robots should adapt to environmental changes and safely interact with humans while performing a diverse range of tasks that conform to sequential and time-critical behaviors.
Existing end-to-end frameworks~\cite{can, code_lang, lifelong} have demonstrated robots performing diverse tasks, while model-based methods~\cite{dynamicallystable, robust} focus on predefined motions that are stable and safe. However, they do not address the following challenges jointly: (i) formal guarantees on safety and satisfying task specifications, and (ii) reactivity to online environmental events and unforeseen disturbances -- which are critical for deploying robots in human-centric environments.
% We address these challenges by combining the reactive task planning capabilities of temporal logic with control theoretic stability and safety guarantees.

Consider the scenario in Fig.~\ref{fig:board-snaps}, where the robot is wiping a whiteboard. We want the robot to switch between periodic wiping motions on the left and right side of the board as commanded by the user. If the camera detects a blue stain, then the robot should wipe that off quickly. If there is an external perturbation where the robot drops the eraser, then it should pick it up and continue the wiping motion. This involves switching between different motions (left and right wiping), reacting to environmental events (stain), time-critical tasks (wiping stain), and human interactions~(eraser~dropped).  

The key challenge for robots in such interactive scenarios is maintaining safety and compliance while reacting to online environmental events and switching between tasks. To alleviate this, \textit{we develop a computationally efficient online planning strategy that generates safe and reactive motions for human-robot interaction, by integrating a 200~Hz discrete planner based on temporal logic and a 1~KHz continuous Dynamical System (DS)-based planner}.

\textbf{Task specifications:} At the task level, one way to describe desired robot behaviors is via temporal logic specifications~\cite{ltl_motion_survey}, previously used in a variety of tasks such as cooking~\cite{temporal_imit}, mobile manipulation~\cite{reactive_yiannis}, search and rescue missions~\cite{GR1}. Such specifications allow robots to understand and interpret commands involving temporal relationships using structured language. For example, a high-level specification for the scenario in Fig.~\ref{fig:board-snaps} can be given as
\vspace{-5pt} 
\begin{align*}
& &&\left(\uncontrollable{left} \to \controllable{wipe left}\right) \ \textnormal{and} \left(\uncontrollable{right} \to \controllable{wipe right}\right) \ \textnormal{and} \\ 
& &&\left(\uncontrollable{blue stain} \to \controllable{wipe stain in 5 seconds}\right) \ \textnormal{and} \\ 
& &&\left(\uncontrollable{no eraser} \to \controllable{pick up eraser}\right).
\numberthis \label{board_example}
\end{align*}
Specifications like~\eqref{board_example} are typically satisfied by integrating high-level discrete actions and low-level continuous motions via formal methods for control~\cite{fully, temporal_dynamic, formal_belta}. Such methods compute a control signal for predefined scenarios but lack reactivity to online events and disturbances. Here, we focus on reactive planning, i.e., the robot should adapt its behaviors depending on the changing environmental observations online. Prior work using Signal Temporal Logic~(STL)~\cite{STL_timing} relies on computationally intensive mixed-integer linear programs~\cite{raman2015reactive, multi_reconfigure} for online adaptation, while methods based on Linear Temporal Logic~(LTL)~\cite{belta_LTL, GR1, reactive_belta, farhad_LTL} require computationally taxing planners on discrete transition systems. To tackle such computational burdens and inflexibility, we instead translate STL specifications into constraints on the continuous motion level enforced via efficient Quadratic Programs (QPs) for real-time planning as in~\cite{stl_lars}.
% We base our task planning approach on event-based STL~\cite{event_STL}, which combines LTL for desired reaction to discrete events in the environment and STL~\cite{stl_lars} for time-critical behaviors

\textbf{Discrete task plan:} We adapt ideas from~\cite{event_STL}, and formally define \controllable{controllable} and \uncontrollable{uncontrollable} propositions in Section~\ref{sec:control_prop} to integrate high level reactive temporal logic specifications with continuous low level motions. Intuitively, \controllable{controllable} propositions, highlighted in \controllable{blue} in~\eqref{board_example}, represent robot behaviors that can always be controlled by appropriate choice of motion planners. \uncontrollable{Uncontrollable} propositions, highlighted in \uncontrollable{red} in~\eqref{board_example}, are discrete environmental events that cannot always be controlled by the robot. 
% For the example in Fig.~\ref{fig:board-snaps} and~\eqref{board_example}, \textit{wipe right}, \textit{wipe left}, \textit{wipe stain in 5 seconds} and \textit{pick up eraser} are controllable, while \textit{left}, \textit{right}, \textit{blue stain} and \textit{no eraser} are uncontrollable.
% We define a reactive temporal logic formula in Section~\ref{sec:reactive_TL} that builds upon controllable and uncontrollable propositions, and construct a B\"uchi automaton~\cite{lindemann2020efficient} using existing tools~\cite{spot_2}. We propose an algorithm in Section~\ref{sec:task_plan} as our discrete task planner that decides the sequence of desired behaviors the robot should follow to always satisfy the task specification, despite environmental changes such as the blue stain and perturbation in Fig~\ref{fig:board-snaps}:~(c) and~(d), respectively.
We define a reactive temporal logic formula in Section~\ref{sec:reactive_TL} that builds upon \controllable{controllable} and \uncontrollable{uncontrollable} propositions. We then construct a B\"uchi automaton~\cite{lindemann2020efficient} using existing tools~\cite{spot_2} and employ it within a discrete task planner that decides the sequence of desired behaviors the robot should follow to always satisfy the task specification, despite environmental changes such as the blue stain and perturbation in Fig~\ref{fig:board-snaps}.
% Section~\ref{sec:task_plan}
% \vspace{-1.5pt}

\textbf{Continuous motion plan:} We integrate our discrete task planner with a continuous DS-based motion planner that executes each desired robot behavior. While our method is agnostic to the underlying DS motion policy, we focus on two types that encapsulate diverse tasks: (i) nominal motions given by an autonomous DS, and (ii) time-critical tasks specified by STL. Nominal motions could either be provided as a library of autonomous DS, or learned from demonstrations~\cite{billard2022learning, temporal_imit}. We utilize our prior work~\cite{nawaz2023learning} that uses Neural ODEs and tools from Lyapunov theory to learn nominal (possibly periodic) motions from only 3-4 demonstrations while also guaranteeing safety and stability. We model time-critical tasks as STL specifications and time-varying Control Barrier Functions~(CBFs)~\cite{stl_lars}. Since our prior work~\cite{nawaz2023learning} also uses CBFs for safety and Control Lyapunov Functions~(CLFs) for stability, we endow robots the capability to follow both time-critical behaviors and complex nominal motions in a single framework by solving an efficient QP online. Thus enabling the robot to react to environmental events and follow the desired behavior, guaranteeing task satisfaction even in the face of disturbances.

%\vspace{-1.5pt}
\textbf{Contributions:} First, we adapt the notion of controllable propositions from~\cite{event_STL} to model more general continuous time robot behaviors such as nominal DS-based motions and time-critical STL tasks. We use such propositions as the basic building blocks of a reactive temporal logic formula to represent task specifications in structured language. Second, we propose a discrete task planning algorithm running online at 200 Hz that decides the sequence of robot behaviors to execute to satisfy the task specification despite environmental changes online. Third, we integrate our task planner with a continuous DS-based motion planner that uses CLFs and CBFs to follow nominal motion plans and satisfy time-critical STL tasks in the presence of unforeseen disturbances, by solving a QP online at 1 KHz. Finally, we validate our approach on the Franka robot arm that performs reactive wiping tasks on a whiteboard and human mannequin using environmental observations from a RealSense camera, while also being compliant to human interactions.   
\vspace{-2.5pt}
\section{Preliminaries}
\label{sec:prelim}
\vspace{-2.5pt}
We introduce the background to define our reactive temporal logic specification, and tools from control theory to satisfy time-critical tasks and guarantee disturbance rejection. Let $x \in \mathcal{X} \subset \mathbb{R}^d$ and $u \in \mathcal{U} \subset \mathbb{R}^m$ be the state and control input for the nonlinear control affine dynamical system
\vspace{-5pt}
\begin{equation}
\dot{x} = f(x) + g(x)u,
    \label{nonlinear_DS}
\end{equation}
where, $f : \mathbb{R}^d \to \mathbb{R}^d$ and $g : \mathbb{R}^d \to \mathbb{R}^{d \times m}$.

\vspace{-3pt}
\subsection{Linear Temporal Logic and B\"uchi Automaton}
\label{sec:ltl_automaton}

We use LTL to specify in structured language the desired reaction of the robot to discrete environmental events. Consider a finite set of atomic propositions $AP$, where each $p \in AP$ can either be \texttt{true} or \texttt{false}. An LTL formula $\varphi$ is constructed using the following syntax.
\vspace{-5pt}
\begin{equation}
\varphi ::= \texttt{true} \ | \ p \ | \ \neg \varphi \ | \ \varphi_1 \land \varphi_2 \ | \ \varphi_1 U \varphi_2 
    \label{ltl_form}
\end{equation}
\vspace{-3pt}
The symbol $::= $ means that $\varphi$ in~\eqref{ltl_form} is assigned to be one of the expressions from the right hand side separated by vertical bars~$|$. The different symbols mean the following: $\neg$ is negation, $\land$ is conjunction and $U$ is the until operator that expresses that $\varphi_1$ should be true until $\varphi_2$ becomes true, where $\varphi_1, \varphi_2$ are well-defined LTL formulas.

The set of discrete environmental observations and robot behaviors that dictate the LTL formula $\varphi$ is given by $AP$. The specification $\varphi$ is a combination of Boolean and temporal operations between the elements of $AP$. For example, consider a task where the robot should add oil to a pan once the pan is hot. The set $AP = \{hot, oil\}$ which corresponds to whether the pan is hot $(hot = 1)$ and oil is poured on the pan $(oil = 1)$. The LTL formula is ${\varphi := G \left(hot \Rightarrow F oil\right)}$, where $G$ denote \textit{always} and $F$ denote \textit{eventually}~\cite{temporal_dynamic}.

An input signal is defined as $\sigma : \mathbb{Z}^+ \to \{0, 1\}^{|AP|}$, which is a discrete time signal that represents the boolean values of the propositions~$AP$. We are interested in signals that satisfy a task specification~$\varphi$. The symbol $(\sigma, t) \vDash \varphi$ indicates that the signal $\sigma$ satisfies a formula $\varphi$ at time $t$. We refer the reader to~\cite{stl_lars} for details on LTL semantics.  

\textbf{B\"uchi automaton}: A B\"uchi automaton~\cite{formal_belta} is a tool that gives A formal way of
deciding whether signals satisfy LTL
specifications. Given $\varphi$, we can construct a deterministic B\"uchi automaton ${\mathcal{B}_{\varphi} = (\mathcal{S}, s_0, AP, \delta, \mathcal{F})}$, where $\mathcal{S}$ is a finite set of states, $s_0 \in \mathcal{S}$ is the initial state, $AP$ is the finite set of atomic propositions, $\delta : \mathcal{S} \times \{0, 1\}^{|AP|} \to \mathcal{S}$ is a transition function such that $s'' = \delta(s', d)$ indicates a transition from $s'$ to $s''$ when the input label is $d \in \{0, 1\}^{|AP|}$, and $\mathcal{F} \subseteq \mathcal{S}$ is a set of accepting states. Given an input signal $\sigma : \mathbb{Z}^+ \to \{0, 1\}^{|AP|}$, a run of the B\"uchi automaton $\mathcal{B}_{\varphi}$ is an infinite state sequence $s:=\{s_0, s_1, s_2, \ldots\}$ that satisfies the transition relation $s_{i+1} = \delta(s_i, \sigma(i))$ for each $i \in \mathbb{Z}^+$. An accepting run on $\mathcal{B}_{\varphi}$ is a sequence $s$ for which inf$(s) \cap \mathcal{F} \neq \emptyset$, where inf$(s)$ is the set of states in the sequence~$s$ that appear infinitely often. If the sequence $s$ is an accepting run on $\mathcal{B}_{\varphi}$, then the input signal~$\sigma$ satisfies~$\varphi$~\cite{formal_belta}. 

\vspace{-2.5pt}
\subsection{Signal Temporal Logic and Control Barrier Functions}
\label{sec:stl}
\vspace{-2.5pt}
We use STL to specify time-critical tasks, time-varying CBFs to satisfy STL tasks and CLFs to enforce stability.

Instead of atomic propositions in LTL, predicates $\mu$ describe truth values in STL using a predicate function~${h(x): \mu = \texttt{true} \Leftrightarrow h(x) \geq 0}$ and $\mu=\texttt{false}$~otherwise,
where $x \in \mathbb{R}^d$ is the robot state. The syntax and semantics of STL closely resemble those of LTL, but can encode explicit quantitative time constraints. We direct the reader to~\cite{stl_lars} for details on STL syntax and semantics.

\textbf{Control Barrier Functions:} We use time-varying CBFs~\cite{stl_lars} to satisfy STL specifications. Satisfying a time-critical task specification is defined as the forward invariance of time-varying safety set $\mathcal{C}(t) \subseteq \mathcal{X} \ \forall \ t \geq 0$ for the system~\eqref{nonlinear_DS}. The set $\mathcal{C}(t)$ is defined as the super-level set of a continuously differentiable function~$B: \mathbb{R}^d \times \mathbb{R}_{\geq 0} \to \mathbb{R}$: $\mathcal{C}(t) = \{x \in \mathcal{X} : B(x, t) \geq 0\}$.
Our objective is to find a control input~$u$ such that the states $x$ that evolve according to the dynamics~\eqref{nonlinear_DS} always stay inside the set $\mathcal{C}(t)$. Such an objective is formalized using \textit{forward invariance} of the set $\mathcal{C}(t)$.
The safe set $\mathcal{C}(t)$ is forward invariant for a given control law $u$ if for every initial point $x(0) \in \mathcal{C}(0)$, the future states $x(t) \in \mathcal{C}(t)$ for all $t \geq 0$. A continuously differentiable function ${B: \mathbb{R}^d \times \mathbb{R}_{\geq 0} \to \mathbb{R}}$ is a \textit{time-varying CBF} for~\eqref{nonlinear_DS} if there exists an extended class $\mathcal{K}_{\infty}$ function $\gamma(\cdot)$ such that for all ${\left(x, t\right) \in \mathcal{X} \times \mathbb{R}_{\geq 0}}$,
\begin{equation}
\sup_{u \in\mathcal{U}} \frac{\partial B(x, t)}{\partial x}^{\top} \left(f(x) + g(x)u\right) +  \frac{\partial B(x, t)}{\partial t}\geq -\gamma(B(x, t))
    \label{CBF_cond}
\end{equation}
The set of all control inputs that satisfy the condition in~\eqref{CBF_cond} for each ${\left(x, t\right) \in \mathcal{X} \times \mathbb{R}_{\geq 0}}$ is $K(x, t) := \{u \in \mathcal{U} : \frac{\partial B(x, t)}{\partial x}^{\top} \left(f(x) + g(x)u\right) + \frac{\partial B(x, t)}{\partial t}\geq -\gamma(B(x, t))\}$. 

The formal result on forward invariance using time-varying CBFs follows from~\cite{stl_lars}.
\begin{theorem}
Let $u(x, t) \in K(x, t)$ be a feedback control law that is Lipschitz in $x$ and piecewise continuous in $t$. Then, the set $\mathcal{C}(t)$ is forward invariant for the control law $u(x, t)$ if $B(x, t)$ is a time-varying control barrier function.
    \label{thm:CBF}
\end{theorem}

STL specifications can be enforced via time-varying CBFs following the construction from~\cite{stl_lars}. For example, consider the STL formula~${\phi ::= F_{[0, 5]}\|x - x^*\| \leq \epsilon}$ which specifies that eventually ($F$), the robot $x$ should be $\epsilon$-close to a target point $x^*$ within the time interval~$[0, 5]$. Similar to~\cite{stl_lars}, we can define a CBF $B(x, t) = \gamma(t) - \|x(t) - x^*\|$, where $\gamma(t)$ defines the rate at which the robot reaches $x^*$ such that $\gamma(5) = \epsilon, \gamma(0) \geq \|x(0) - x^*\|$. 

We use time-independent CBFs and CLFs to guarantee safety and stability of nominal motion plans given by autonomous DS models, which is detailed in prior work~\cite{ames2019control, nawaz2023learning}. The difference between a time-independent CBF defined in~\cite{ames2019control} and the time-varying CBF is the presence of the term $ \frac{\partial B(x, t)}{\partial t}$ in~\eqref{CBF_cond}, which makes constraint satisfaction more difficult if $\frac{\partial B(x, t)}{\partial t} < 0$. A \textit{Control Lyapunov Function~(CLF)}~${V : \mathbb{R}^d \to \mathbb{R}_{\geq 0}}$ is a special case of a time-independent CBF~$B : \mathbb{R}^d \to \mathbb{R}$ with $B(\cdot) = -V(\cdot)$ and the time-independent safe set $\mathcal{C} = \{0\}$, where Theorem~\ref{thm:CBF} guarantees asymptotic stability: see~\cite{ames2019control} for more details.
\vspace{-2.5pt}
\section{Problem formulation}
\label{sec:probl_form}
We first define a reactive temporal logic formula to describe the task specification using structured language, and formally write the problem statement.  
\vspace{-2.5pt}
\subsection{Controllable Propositions}
\label{sec:control_prop}\vspace{-2.5pt}
We introduce a general notion of controllable propositions that is adapted from~\cite{event_STL} to model discrete events in our task specification that depend only on the state signal~$x(t)$. Prior work~\cite{event_STL} model only STL tasks, but we also focus on autonomous DS-based motions and describe how observations from the continuous time state signal~$x$ translate to a discrete time signal~$\sigma$ of LTL in Section~\ref{sec:ltl_automaton}.     

We define a generalized version of the predicates~$\mu$ described in~\ref{sec:stl} as controllable propositions.
\begin{definition}
The truth value of a controllable proposition~$p_c$ is defined below using a tuple $\left(\mathcal{L}, \mathcal{M}, 
\mathcal{Q}\right)$, where ${\mathcal{L} : \mathcal{P} \times \mathbb{R}_{\geq 0} \to \mathcal{Q}}$ is an operator from the space of state signals~$\mathcal{P} = \{x \ | \ x : \mathbb{R}_{\geq 0} \to \mathbb{R}^d\}$ and time~$\mathbb{R}_{\geq 0}$ to a space of signals $\mathcal{Q} = \{y \ | \ y : \mathbb{R}_{\geq 0} \to \mathbb{R}^q\}$ in cartesian space~$\mathbb{R}^q$, ${\mathcal{M}: \mathcal{P} \times \mathbb{R}_{\geq 0} \to 2^{\mathcal{Q}}}$, where $2^{\mathcal{Q}}$ is the power set of $\mathcal{Q}$.
\vspace{-3pt}
\begin{equation}
p_c = \begin{cases}
    \texttt{true} & \Leftrightarrow \mathcal{L}(x, t) \in \mathcal{M}(x, t)\\
        \texttt{false} & \Leftrightarrow \mathcal{L}(x, t) \notin \mathcal{M}(x, t)
\end{cases}
    \label{control_prop_eqn}
\end{equation}
    \label{control_prop_def}
\end{definition}
\vspace{-5pt}

Intuitively, the operator~$\mathcal{L}$ describes the current behavior of the robot relevant to $p_c$ at time $t$, $\mathcal{M}$ describes the desired behavior relevant to $p_c$ at time $t$, and $\mathcal{Q}$ is the space of signals that is applicable for the controllable proposition~$p_c$. For the example in Fig.~\ref{fig:board-snaps}, we represent all motions in 3D~(${d = 3}$) and wiping motions as autonomous DS~${\dot{x}(t) = f(x(t))}$ so that $q = d$, $\mathcal{L}(x, t) = \dot{x}$ for all ${t \geq 0}$ and ${\mathcal{M}(x, t) = \{y \ | \ y:\mathbb{R}_{\geq 0} \to \mathbb{R}^3 \ \textnormal{s.t.} \ y(t) = f(x(t))\}}$.   

We can model STL tasks using controllable propositions. First, for the predicates~$\mu$ defined in Section~\ref{sec:stl}, ${q=1,\, \mathcal{L}(x, t) = h \circ x}$ for all $t \geq 0$ and ${\mathcal{M}(x, t) =  \{y \ | \ y:\mathbb{R}_{\geq 0} \to \mathbb{R} \ \textnormal{s.t.} \ y(t) \geq 0\}}$ for all ${x \in \mathcal{P}}$. Then, an STL specification ${F_{[0, 5]} \|x - x^*\| \leq \epsilon}$ can be modeled as ${\mathcal{L}(x, t) = h \circ x}$ for all ${t \geq 0}$, where ${h(x(t)) = -\|x(t) - x^*\| + \epsilon}$, $q=1$ and ${\mathcal{M}(x, t) = \{y \ | \ \exists \ t' \in [t, t+5] \ \textnormal{s.t.} \ y(t') \geq 0 \}}$ ~$\forall {x \in \mathcal{P}}$. 

% Following a nominal autonomous dynamical system~${\dot{x} = d(x)}$ for a time period $[t, t + T]$ is also a controllable proposition, where $t$ is the current time, ${q=d, \ \mathcal{L}(x, t) = \dot{x}}$ and $\mathcal{M}(x, t) =  \{y \ | \  {y : \mathbb{R}_{\geq 0} \to \mathbb{R}^d} \ \textnormal{s.t.} \ y(t') = d(x(t')) \ \forall \ t' \in [t, t+T]\}$. We switch between different autonomous DS online at task execution which correspond to satisfying different controllable propositions at each time step. 
We denote the propositions $p_c$ in Definition~\ref{control_prop_def} as \textit{controllable} because their truth value entirely depends on the state signal~$x$ which can be controlled by $u$ in~\eqref{nonlinear_DS}, but we do not state any formal equivalence to controllability~\cite{modern_control}. While the notion of controllable propositions is in general abstract, we focus on nominal DS-based motions and STL tasks. 

\textbf{Continuous to discrete time:} We consider a finite set of controllable propositions $\mathcal{A}_c$, where each $p_c \in \mathcal{A}_c$ can either be \texttt{true} or \texttt{false} as defined by~\eqref{control_prop_eqn}. An LTL formula can be defined with appropriate syntax and semantics as given in Section~\ref{sec:ltl_automaton} with ${AP = \mathcal{A}_c}$. However, the signal~$\sigma$ in Section~\ref{sec:ltl_automaton} that dictate the boolean values of $AP$ is a discrete time signal, but controllable propositions depend on the continuous time state signal $x$. In practice, we fully observe the state $x(t)$ at each continuous time $t \geq 0$, but update the boolean values of the controllable propositions only at discrete time steps~$k \in \mathbb{Z}^+$. We connect the different time scales by denoting~$\{t_0, t_1, \ldots, \}$ to be a possibly irregular sampling of the continuous time interval~$[0, \infty)$ with $t_0 = 0$, where each $t_k$ corresponds to the discrete time step~$k$. We define a \textit{controllable signal}~${\sigma_c : \mathbb{Z}^+ \to \{0, 1\}^{\left|\mathcal{A}_c\right|}}$ such that for all $k \in \mathbb{Z}^+$, $\sigma_c(k)$ depends on the state signal~$x$ and time $t_k$. Now we can use a B\"uchi automaton as described in~\ref{sec:ltl_automaton} for any signal ${\sigma _c: \mathbb{Z}^+ \to \{0, 1\}^{|\mathcal{A}_c|}}$ to verify satisfiability of LTL formulae over $\mathcal{A}_c$.
\vspace{-2.5pt}
\subsection{Reactive temporal logic}
\label{sec:reactive_TL}
\vspace{-2.5pt}
We define the syntax of a \textit{Reactive Temporal Logic}~(RTL) formula that the robot should satisfy, using LTL formulae over controllable and uncontrollable propositions. Such a formula describes how the robot should react to environmental events by following different desired behaviors. In addition to $\mathcal{A}_c$ described in~\ref{sec:control_prop}, we define \textit{uncontrollable propositions}~$\mathcal{A}_u$ that represent discrete environmental observations which cannot always be controlled by the robot, such as \textit{left, right, blue stain, no eraser} described in Section~\ref{sec:intro}. The truth values of~$\mathcal{A}_u$ are determined by both the state signal~$x$ and an observed \textit{uncontrollable signal}~$\sigma_u : \mathbb{Z}^+ \to \{0, 1\}^{|\mathcal{A}_u|}$. The syntax of RTL formulas $\Psi$ is
\vspace{-3pt}
\begin{equation}
\Psi ::= \psi \ | \ G (\varphi \Rightarrow \Psi) \ | \ G (\psi \Rightarrow \Psi) \ | \ \Psi_1 \land \Psi_2
    \label{reactive_TL_syntax}
\end{equation}
where, $\psi$ is an LTL formula defined in~\eqref{ltl_form} with $AP = \mathcal{A}_c$, $\varphi$ is an LTL formula with $AP = \mathcal{A}_u$, and $G$ indicates the always operator~\cite{temporal_dynamic}. The semantics of $\Psi$ follow that of LTL in Section~\ref{sec:ltl_automaton} with $AP = \mathcal{A}_c \cup \mathcal{A}_u$ and $\mathcal{A}_c \cap \mathcal{A}_u = \emptyset$. We can construct B\"uchi Automaton $\mathcal{B}_{\Psi}$ to verify if an input signal $\sigma : \mathbb{Z}^+ \to \{0, 1\}^{|AP|}$ satisfies $\Psi$. For example, consider a simple reactive cooking task where the robot should keep stirring a pan if it is hot. Otherwise, the robot should increase the temperature of the stove by pressing a button. The RTL specification would be ${\Psi := G (\textit{hot} \Rightarrow \textit{stir}) \land G (\neg \textit{hot} \Rightarrow F \textit{(press button)})}$, where ${\mathcal{A}_u = \{hot\}, \mathcal{A}_c = \{stir, \textit{press button}\}}$. Note that the syntax in~\eqref{reactive_TL_syntax} does not allow formulas such as $G(stir \Rightarrow hot)$ since $hot \in \mathcal{A}_u$. Such a formula does not make sense for our work since we always focus on reacting to environmental observations represented by $\mathcal{A}_u$ by following a desired behavior dictated by $\mathcal{A}_c$. 
\vspace{-2.5pt}
\subsection{Task and Motion Plan}
\label{sec:task_motion_plan}
Given a set of nominal autonomous DS and an RTL specification~$\Psi$, we aim to develop a reactive planning strategy to satisfy $\Psi$ while being adaptive to environmental events and stable  with respect to disturbances. We use a DS-based motion plan for a robotic manipulator defined as
\vspace{-3pt}
\begin{equation}
    \dot{x} = \hat{f}_{n(t)}(x) + u(x, t),
\label{DS_form}
\vspace{-3pt}
\end{equation}
where, $\hat{f}_{n(t)}(x)$ encodes a nominal motion~${n(t) \in \{1,2,\ldots,N\}}$ at time $t$, while $u(x, t)$ is an auxiliary control signal used to satisfy the task specification and enforce disturbance rejection properties. We assume that the $N$ autonomous dynamical systems~${\mathcal{D} = \{\hat{f}_i\}_{i=1}^N}$ that govern the nominal motions are given to us; these may, for example, be learned from demonstrations~\cite{nawaz2023learning}. We choose~$u(x, t)$ and~$n(t)$ for the robot to follow at each time step based on the discrete controllable signal~$\sigma_c$ that satisfy $\Psi$, which further depends on the uncontrollable signal~$\sigma_u$. Hence, we have the below assumption.
\begin{assumption}
For all ${\sigma_u : \mathbb{Z}^+ \to \{0, 1\}^{\left|\mathcal{A}_u\right|}}$, there exists a~${\sigma_c : \mathbb{Z}^+ \to \{0, 1\}^{\left|\mathcal{A}_c\right|}}$ that satisfies a given RTL specification~$\Psi$ where at most one controllable proposition~${\sigma_c(k) \in \mathcal{A}_c}$ is \texttt{true} at each time step~$t_k \geq 0$.  
\label{assum_1}
\end{assumption}

The above assumption asserts the existence of at-least one controllable signal~$\sigma_c$ that satisfies~$\Psi$, even if the uncontrollable propositions change online. Without Assumption~\ref{assum_1}, there might exist a~$\sigma_u$ where no~$\sigma_c$ can satisfy~$\Psi$. Such scenarios are known as deadlock modes~\cite{deadlock}, where a supervisory signal~\cite{supervisory} is required to recover the system, which we reserve for future work. We model controllable propositions as following low level continuous motions and assume that the robot can follow at-most one motion at a given time, since we allow switching online between different motions. 

We are given an RTL specification $\Psi$ satisfying Assumption~\ref{assum_1} and a set of nominal autonomous dynamical systems $\mathcal{D}$ specifying a DS-based family of motion plans~\eqref{DS_form}. Our task is formalized in the following problem statement: 

\textit{Develop a reactive control strategy such that at each discrete time~$t_k$, and for any uncontrollable signal~$\sigma_u$ the solution~$x(t)$ to~\eqref{DS_form} generates a controllable signal~$\{\sigma_c(k), \sigma_c(k+1), \ldots\}$ that always satisfy $\Psi$.}

% \textit{Develop a reactive control strategy such that at each discrete time~$t_k$, and for any uncontrollable signal~$\sigma_u$, there exists a continuous time feedback motion policy~$u(x, t)$ and nominal motion plan choice~$n(t)$ so that the solution~$x(t)$ to~\eqref{DS_form} generates a controllable signal~$\{\sigma_c(k), \sigma_c(k+1), \ldots\}$ that always satisfy $\Psi$.}

The above problem statement requires the robot to react online to uncontrollable environmental changes given by $\sigma_u$ while also adapt to unforeseen external disturbances via the auxiliary input~$u(x, t)$.
\vspace{-2pt}
\section{Proposed Methodology}
\label{sec:approach}
\vspace{-3pt}
We present our modular control architecture that integrates a discrete task planner to satisfy the specification $\Psi$ and a continuous DS-based motion planner to reject external disturbances while adapting to dynamic environmental changes. Unlike prior approaches~\cite{lifelong, dynamicallystable, robust} that do not jointly show reactivity to online environmental events and stable task execution, we propose a single framework that demonstrates reactive robot tasks with formal stability guarantees for physical robot interaction. We will use CLFs and CBFs described in Section~\ref{sec:stl} for our parameterization~\eqref{DS_form}, which is a control-affine dynamical system~\eqref{nonlinear_DS} for the time period when $n(t)$ is constant with $f(x) := \hat{f}_{n(t)}(x)$ and $g(x) := I$. 

\begin{figure}[!t]
         \centering
         \includegraphics[width=0.45\textwidth]{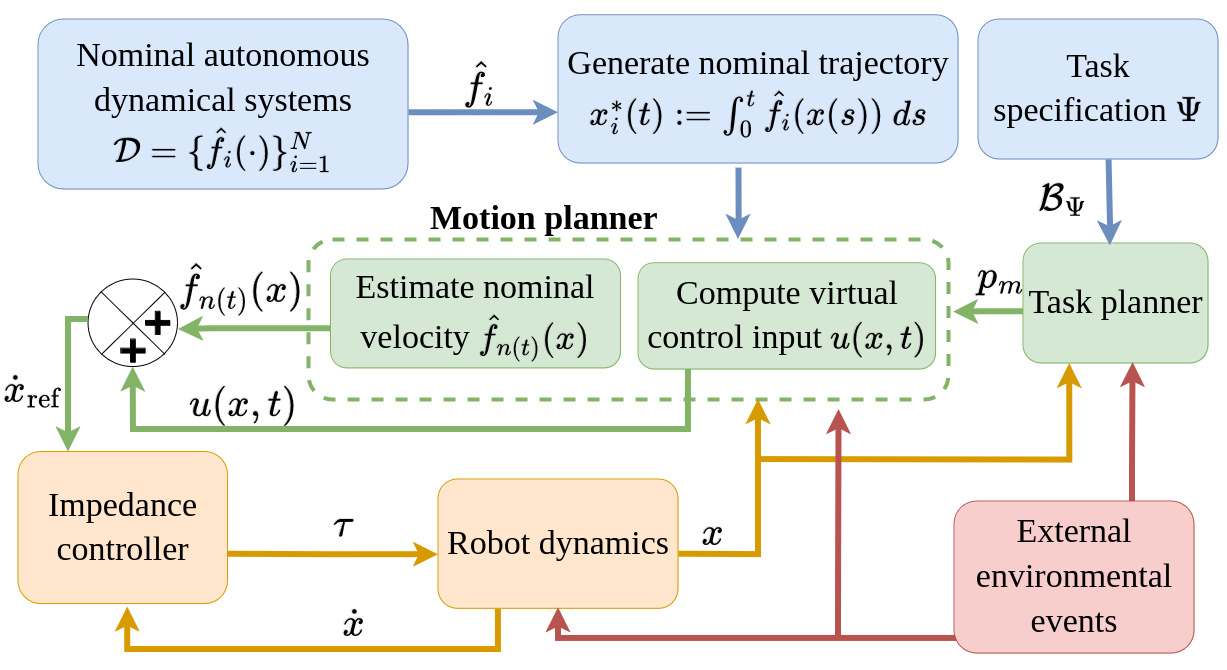}
      \caption{\small The control flow of our proposed pipeline. The state of the robot is $x$, the low level control input are the joint torques $\tau$, and the desired velocity in state space is $\dot{x}_\mathrm{ref}$.}
        \label{fig:block_diag}
\end{figure}

A schematic of the proposed control flow is presented in Fig.~\ref{fig:block_diag}. The blue blocks represent the offline components and the green blocks are the online planning modules. The autonomous DS  models~$\hat{f}_i(x)$ could either be given by a library of motion primitives~\cite{DMP}, or can be learned from demonstrations~\cite{billard2022learning, nawaz2023learning}. The user specifies a reactive temporal specification $\Psi$ that the robot should satisfy and we construct the corresponding B\"uchi automaton $\mathcal{B}_{\Psi}$ using automated tools~\cite{spot_2}. At deployment, given an observation of the external environmental events, our task planner decides the type of robot behavior~$p_m \in \mathcal{A}_c$ to follow at each time step using $\mathcal{B}_{\Psi}$ so that $\Psi$ is satisfied. The details of the task planner on how to decide~$p_m$ are given in Section~\ref{sec:task_plan}. Then, the motion planner computes the reference velocity $\dot{x}_{\mathrm{ref}}$ to follow~$p_m$ based on~\eqref{DS_form} using the \textit{virtual control input}~$u(x, t)$, which enforces disturbance rejection properties and satisfies STL specifications included in $\Psi$. We denote the reference velocity for the low level controller as $\dot{x}_{\mathrm{ref}}$, which in general may be different from the real velocity $\dot{x}$ of the robot. The reference velocity $\dot{x}_{\mathrm{ref}}$ is given as input to an impedance controller~\cite{impedance} that computes the low level control input~$\tau$ (joint torques) for the physical robotic system.  We emphasize that the virtual control input~$u(x, t)$ is different from the low-level control inputs~$\tau$ given in Fig.~\ref{fig:block_diag}, and~$u(x, t)$ is a component of the motion planning DS~\eqref{DS_form}.
% \begin{wrapfigure}{R}{0.5\textwidth}
\RestyleAlgo{ruled}
\LinesNumbered
%% This is needed if you want to add comments in
%% your algorithm with \Comment
% \SetKwComment{Comment}{/* }{ */}
\setlength{\textfloatsep}{0pt}% Remove \textfloatsep
% \addtocounter{algocf}{1}
\begin{algorithm}[!b]
\caption{Task Planner}
\label{alg:task_plan}
\small
\KwInit{\\
$\mathcal{B}_{\Psi}:=(\mathcal{S}, s_0, \mathcal{A}_c \cup \mathcal{A}_u, \delta, \mathcal{F}) \gets \textnormal{Automaton}(\Psi)$\;
$s' \gets s_0, \ k \gets 0$\;
}
\While{task running}{
$[\sigma_c(k), \sigma_u(k)] \gets \textnormal{UpdateSensor()}$\;
  $s \gets \delta(s', [\sigma_c(k), \sigma_u(k)])$\;
\eIf{$k=0 \ \textnormal{or} \ \sigma_u(k) \neq \sigma_u(k-1)$}{
$j \gets 0$\;
  $\mathcal{G}(\mathcal{S}, E) \gets \textnormal{Graph}(\mathcal{B}_{\Psi}, \sigma_u(k))$\;
    $\{s_0, s_1, s_2, \ldots \} \gets \textnormal{ShortestPath}(\mathcal{G}, s_0, \mathcal{F})$\;
  }{
  \If{$s \neq s'$}{
  $j \gets j+1$\;
    }
    }
$p_m \gets p_c(k)$, where $p_c(k) \in \mathcal{A}_c$ such that $p_c^j[p_c(k)] \vDash \texttt{true}$ and $\delta(s_j, [p_c^j, \sigma_u(k)]) = s_{j+1}$\;
  $s' \gets s, k \gets k + 1$\;
  $\textnormal{MotionPlan}(p_m)$\;}
\end{algorithm}
\vspace{-1pt}
\subsection{Task Planner}
\label{sec:task_plan}
\vspace{-2.5pt}
We propose Algorithm~\ref{alg:task_plan} that outputs the behavior~$p_m$ the robot should follow at each time step to satisfy~$\Psi$, even when the environmental events are dynamically changing.

\textbf{Description of Algorithm~\ref{alg:task_plan}}: Given a task specification $\Psi$, we construct the B\"uchi automaton ${\mathcal{B}_{\Psi}= (\mathcal{S}, s_0, AP, \delta, \mathcal{F})}$ using Spot~2.0~\cite{spot_2}. We denote $s$ and $s'$ to be the current and previous state of the automaton, respectively, and $k$ is the current discrete time step. Lines 2-15 are implemented online at 200~Hz in the task planner block of Fig.~\ref{fig:block_diag}. In line~6, we form a directed graph~$\mathcal{G}(\mathcal{S}, E)$ from~$\mathcal{B}_{\Psi}$, which is implemented only at $k=0$ or if there is a change in the values of the uncontrollable propositions~$\sigma_u(k)$. The nodes of $\mathcal{G}$ is $\mathcal{S}$, while the edges $E$ reflect valid transitions on the automaton that comply with the current value of uncontrollable propositions~$\sigma_u(k)$, i.e., for all $s_1, s_2 \in \mathcal{S}$, $(s_1, s_2) \in E$ if and only if there exists a $p \in \{0, 1\}^{|\mathcal{A}_c|}$ such that $\delta(s_1, [p, \sigma_u(k)]) = s_2$. In line~6, we plan a shortest path on $\mathcal{G}(\mathcal{S}, E)$ using breadth-first search~\cite{BFS} that reaches an accepting state in~$\mathcal{F}$ and stays there infinitely often. Such a path always satisfies~$\Psi$ as described in Section~\ref{sec:ltl_automaton}. If there is no change in the values of uncontrollable propositions, we implement lines 9-11 so that the robot moves along the path, where index~$j$ represents the progress along the path to satisfy~$\Psi$. Finally, in line~13,  we decide the type of motion plan~$p_m$ to be implemented. 
% by the motion planner as detailed next.
% based on the transitions of the shortest path, which is detailed below.

\begin{figure}[!t]
         \centering         \includegraphics[width=0.33\textwidth]{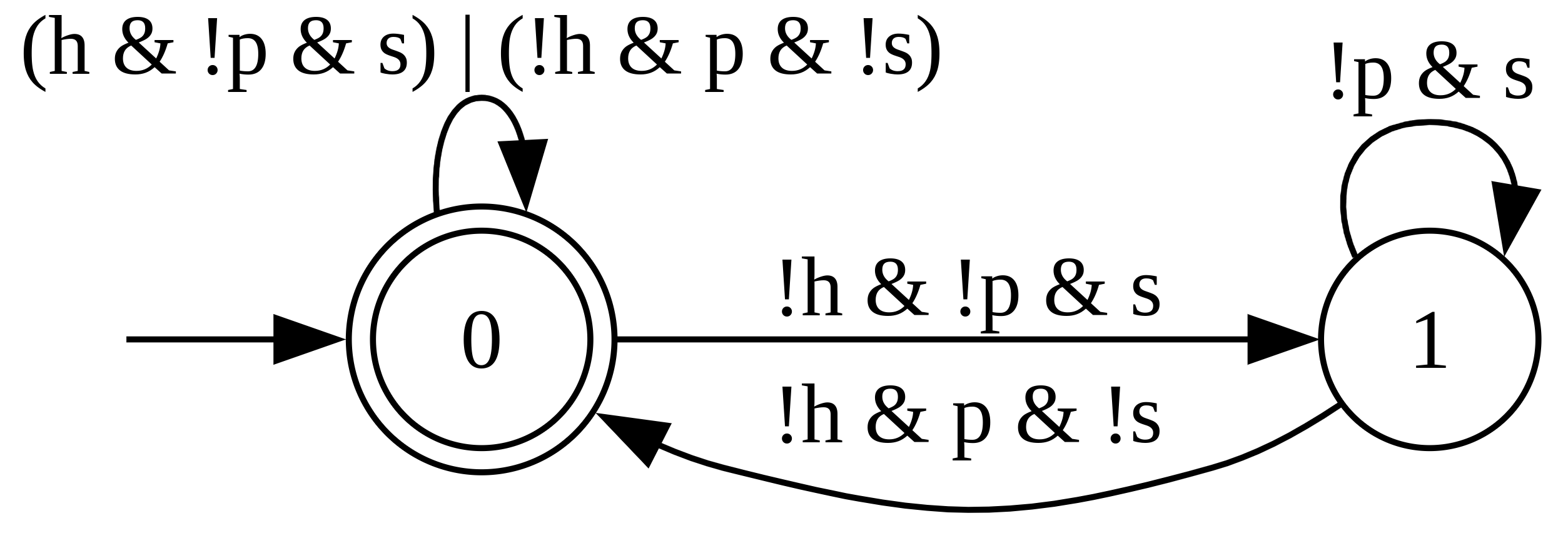}
      \caption{\small Automaton $\mathcal{B}_{\Psi}$ for ${\Psi := G (\textit{h} \Rightarrow \textit{s}) \land G (\neg \textit{h} \Rightarrow F \textit{(p)})}$, where $\textnormal{!p} \equiv \neg \textnormal{p}$.} 
        \label{fig:aut_eg_small}
\end{figure}

\textbf{Decide motion plan~$p_m$:} Let the shortest path obtained in line~6 be $w = \{s_0, s_1, s_2, \ldots\}$, where ${s_0 = s, \ \delta(s_j, [p_c^j, \sigma_u(k)]) = s_{j+1}}$ for all $j \in \mathbb{Z}^+$, and ${p_c^j \in \{0, 1\}^{|\mathcal{A}_c|}}$ is the value of controllable propositions when transitioning from $s_j$ to $s_{j+1}$. Although there could be multiple such $p_c^j$, we observed from our experience that the transitions on the shortest path will have exactly one~$p_c^j$ for each unique~$\sigma_u(k)$.  By Assumption~\ref{assum_1}, for all possible $\sigma_u$ and each ${j \in \mathbb{Z}^+}$, there exists a $p_c^j$ and at most one controllable proposition is \texttt{true} in~$p_c^j$ that satisfies~$\Psi$. Hence, we can always find a path that satisfies~$\Psi$ for all possible environmental changes~$\sigma_u$, i.e., ${\inf\left(w\right) \cap \mathcal{F} \neq \emptyset}$. In line~13, we denote~$p_c(k)\in\mathcal{A}_c$ to be the controllable proposition that is \texttt{true} in~$p_c^j$, where $p_c^j$ is the label of the transition along the path that satisfies~$\Psi$. We thus choose the type of motion plan $p_m$ to be the current controllable proposition~$p_c(k)$ that has to be true to move along the shortest path. For example, the automaton $\mathcal{B}_{\Psi}$ is given in Fig.~\ref{fig:aut_eg_small} for the specification ${\Psi := G (\textit{h} \Rightarrow \textit{s}) \land G (\neg \textit{h} \Rightarrow F \textit{(p)})}$ described in Section~\ref{sec:reactive_TL}, where ${h := hot, \ s:=stir, \ p:=press \ button}$. If the initialization is such that $s_0 = 0$; $h = 1, s = 1, p = 0$, then the shortest path is just the self-transition $0 \to 0$. Then, from the label of the transition $0 \to 0$ and $\sigma_u(k) = \{h, \neg p\}$, the motion plan to be implemented is $p_m = s$ so that the robot follows the behavior \textit{stir}. If $h=0$ while the robot is stirring~$(s = 1)$, then the system transitions to state $1$. Now, the shortest path is computed again and the path is $1 \to 0 \to 0$. From the label of the transitions $1 \to 0 \to 0$ and $h = 0$, the motion plan to be implemented is $p_m= p$ so that the robot follows the behavior \textit{press button}. In the next section, we describe how the desired behavior $p_m$ is achieved by implementing a DS-based motion plan.  
% If there are multiple such $p_c^j$, then we choose them at random since any of them will transition along the shortest path to the accepting states. However, we observed from our experience that the transitions will have exactly one~$p_c^j$ for each unique~$\sigma_u(k)$. 
% If our DS-based motion planner computes a control signal~$u(x, t)$ and chooses~$\hat{f}_{n(t)}(x)$ such that the solution~$x(t)$ to~\ref{DS_form} satisfies the sequence~$\{p_c(k), p_c(k+1), \ldots\}$, then the robot always satisfies the task specification~$\Psi$.  \change{A proposition?} 
\vspace{-2.5pt}
\subsection{Motion Planner}
\label{sec:motion_plan}
\vspace{-2.5pt}
Given the output~$p_m$ of the discrete task planner at each time step~$k$, we solve a QP online at 1 KHz that (i) guarantees satisfaction of STL specifications, and (ii) enforces safety and stability properties while following nominal motions given by autonomous DS. We also propose a simple yet efficient switching strategy so that robots switch between different behaviors when reacting to environmental observations, unlike previous approaches~\cite{raman2015reactive, multi_reconfigure} that solve expensive optimization problems to re-plan. 

\subsubsection{Nominal motion plans}
\label{sec:nominal_motion}
We compute safe and stable motions that converge to a nominal target trajectory which is generated from an autonomous dynamical system. Although one could assume that the set of nominal dynamical systems~$\mathcal{D}$ are given to the user from an existing library of motion primitives~\cite{DMP}, we use the CLF-CBF-NODE from our prior work~\cite{nawaz2023learning} that learns from demonstrations DS models~$\hat{f}_i$ for each behavior~$i \in \{1,2,\ldots,N\}$ that includes complex, possibly periodic motions. For example, the \textit{stir} behavior mentioned in the previous section can be represented by a periodic trajectory~$x^*_1(t)$ using a DS model~$\dot{x} = \hat{f}_1(x)$ that is learned from demonstrations. 

\textbf{CLF-CBF-QP}: We abuse notation and use $x^*(t)$ and $\hat{f}(t)$ to denote the target trajectory and the autonomous DS for a single nominal motion, but the method is the same for any other nominal motion. Let the error be $e(t) = x(t) - x^*(t)$, and for ease of notation, we drop the explicit dependence on time~$t$, and write $e$, $x$, and $x^*$ for the current error, state, and target point at time $t$, respectively. The control-affine error dynamics is $\dot{e} = \hat{f}(x) - \dot{x}^* + u(x)$ from~\eqref{DS_form}. The explicit dependence of $u(x)$ and $\hat{f}(x)$ on time is removed since we focus only on time-independent CBFs to model safety sets, and CLFs for the time period when $n(t)$ is constant. In the next section, we describe how we switch between different behaviors. Then, by Theorem~\ref{thm:CBF}, if there exists a CLF~$V$ for the dynamics $\dot{e}$, then, any feedback control law $u(x) \in K(x)$ will drive the error asymptotically to zero. Similarly, if there exists a time-independent CBF~$B$ corresponding to a safety set~$\mathcal{C} \subseteq \mathcal{X}$ for the dynamics~\eqref{DS_form}, then, any feedback control law $u(x) \in K(x)$ will render the system safe. We combine both the properties of safety and stability and solve the below QP to compute $u(x)$, where we prioritize safety over stability.

Given the current state of the robot~$x$ and the target point~$x^*$, the QP that guarantees a safe motion plan is
\begin{equation}
    \begin{aligned}
\vspace{-10pt}
        \quad & (u(x), \_) = \argmin_{\{v, \eta\}} \quad \bigl\|v\bigr\|^2_2 + \lambda \eta^2 \\
        \textnormal{s.t.} \quad & \frac{\partial B(x)}{\partial x}^{\top}\left(\hat{f}(x) + v\right) \geq -\gamma(B(x))\\
        \vspace{-10pt}
        \quad & \frac{\partial V(e)}{\partial e}^{\top}\left(\hat{f}(x)  - \dot{x}^* + v\right) \leq -\alpha(V(e)) + \eta,
    \end{aligned}
    \label{safety_opt}
\end{equation}
where $\alpha$ is a class $\mathcal{K}$ function, $\eta$ is a relaxation variable to ensure feasibility of~\eqref{safety_opt} and is penalized by $\lambda > 0$. We use Algorithm~1 from our prior work~\cite{nawaz2023learning} to choose~$x^*$ and~$\dot{x}^* = \hat{f}(x^*)$ that ensures tracking of complex periodic trajectories in-spite of external disturbances. If the safety specifications do not disturb the nominal motion, then $\eta = 0$ in~\eqref{safety_opt}, which guarantees stability.

% NF: Alternative for Wiping board figure and table
\begin{figure*}[!b]
  \begin{minipage}{.40\textwidth}
  \vspace{-10pt}
    \centering
    \scriptsize
        \resizebox{\textwidth}{!}{\begin{tabular}{|p{0.4cm}|p{2.60cm}|p{4.2785cm}|}
            \hline
            \textbf{Sym.} & \textbf{Boolean condition} & \textbf{Description} \\
            \hline
            $w_a$ & $w_a = 1 \Leftrightarrow \dot{x}(t) = \hat{f}_a(x(t))$ & Wiping motion by nominal autonomous DS~$\hat{f}_a$, $a \in \{\textnormal{left}, \textnormal{right}\}$ indicates the side on the board.  \\
             \hline
            $w_{\textnormal{stain}}$ & $w_{\textnormal{stain}} = 1$ $\Leftrightarrow \dot{x}(t)=\hat{f}_{\textnormal{stain}}(x(t))$  & Motion to wipe off the detected stain online given by a nominal autonomous DS~$\hat{f}_{\textnormal{stain}}$ \\
            \hline
            $s_p$ & 
            $s_p = 1 \Leftrightarrow F_{[0, t_s]} {\|x(t) - x_s\|} \leq \epsilon$  
            & The end-effector should~eventually $(F)$ be $\epsilon$ close to the stain~$x_s$ within time interval~$[0, t_s]$. \\            
            \hline            
                        $e_p$ & $e_p = 1 \Leftrightarrow {F_{[0, t_e]}\|x(t) - x_e\| \leq \epsilon}$ & The end-effector should eventually be $\epsilon$ close to the eraser~$e_p$ within~$[0, t_e]$. \\
                        \hline
        \end{tabular}}
        \captionof{table}{Controllable propositions (Whiteboard)}\label{tab:controllable_board}
  \end{minipage}
    \begin{minipage}{0.60\textwidth}
    % \vspace{-10pt}
     \includegraphics[width=\textwidth]{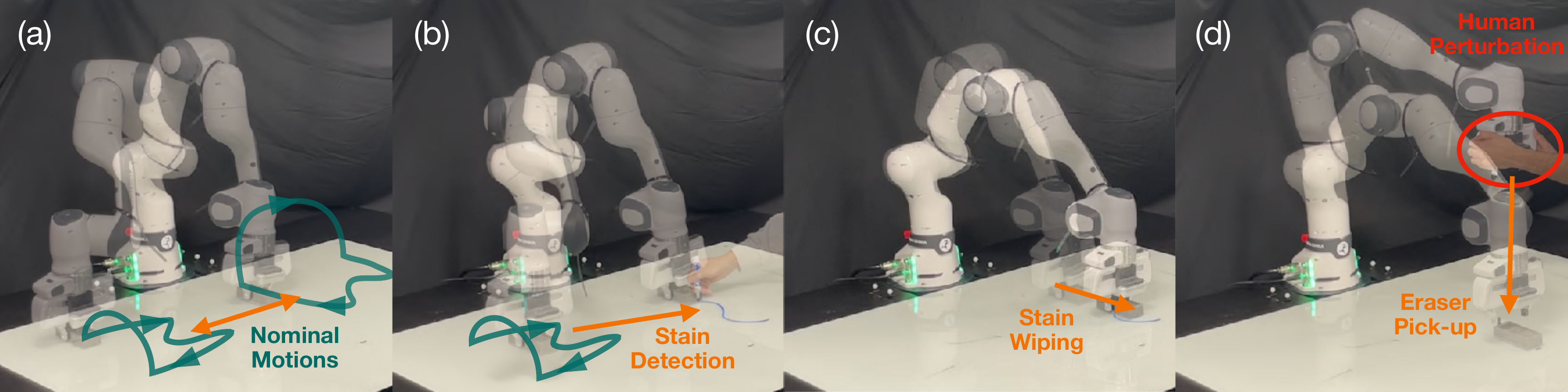}
      \caption{Whiteboard wiping: (a) switching between nominal motions (b) going to the stain position (c) wiping off stain (d) human perturbation and eraser pick up.} 
        \label{fig:board_snaps_exp}
  \end{minipage} 
\end{figure*}

% NF: Alternative for Wiping mannnequin figure and table
\begin{figure*}[!b]
\vspace{-5pt}
  \begin{minipage}{.260\textwidth}
  % \vspace{-10pt}
    \centering
    \scriptsize
        \resizebox{\textwidth}{!}{\begin{tabular}{|p{0.7cm}|p{2.5cm}|p{3cm}|}
            \hline
            \textbf{Symbol} & \textbf{Boolean condition} & \textbf{Description} \\
            \hline
            $w_a$ & $w_a = 1 \Leftrightarrow {\dot{x}(t) = \hat{f}_a(x(t))}$ & Wiping motion on $a \in \{\textnormal{leg, hand, back}\}$ given by a nominal autonomous DS~$\hat{f}_a$ \\
            \hline
            $d_p$ & $d_p = 1 \Leftrightarrow {F_{[0, t_d]} \|x(t) - x_d\| \leq \epsilon}$  & The end-effector should eventually $(F)$ be $\epsilon$-close to the dip position~$x_d$ within the time interval~$[0, t_d]$. \\            
            \hline            
                        $b_p$ & $b_p = 1 \Leftrightarrow {F_{[0, t_b]}\|x(t) - x_b\| \leq \epsilon}$ & The end-effector should eventually be $\epsilon$-close to the bowl position~$x_b$ within~$[0, t_b]$. \\
                        \hline
        \end{tabular}}
        \captionof{table}{Controllable propositions (mannequin)}\label{tab:controllable_m}
  \end{minipage}
  \vspace{-15pt}
    \begin{minipage}{0.74\textwidth}
     \includegraphics[width=\textwidth]{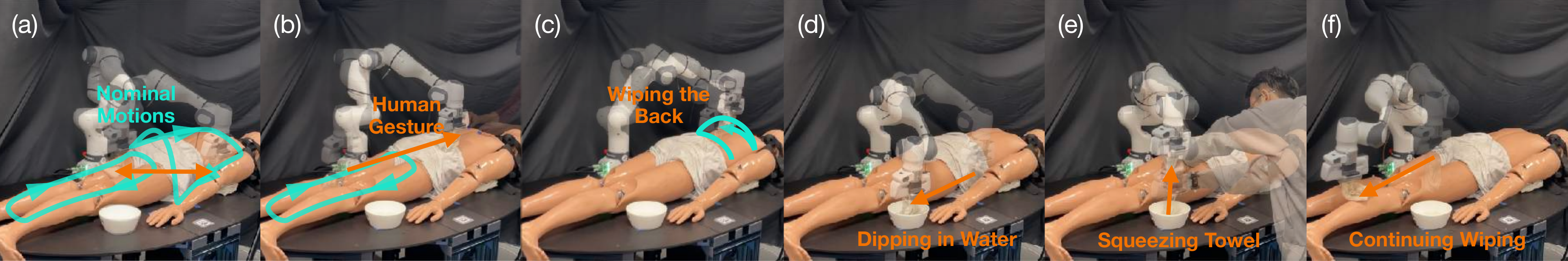}
      \caption{Mannequin wiping: (a) switching between nominal motions (b) going to the back (c) wiping the back (d) dipping the towel in water (e) human squeezing the wet towel and (f) continuing wiping.} 
        \label{fig:mannequin_snaps}
  \end{minipage}  
\end{figure*}

\subsubsection{STL tasks}
\label{sec:CBF_STL}
We use time-varying control barrier functions to satisfy time-critical tasks specified by STL~\cite{stl_lars}. For example, consider the STL task ${\phi ::= F_{[a, b]}\|x - x^*\| \leq \epsilon}$ which encodes that the robot's end-effector position $x$ should eventually be $\epsilon$ close to a goal point $x^*$ within the time interval~$[a, b]$. Such a task $\phi$ could represent the \textit{press button} behavior mentioned in Section~\ref{sec:task_plan}, where the robot should be in close proximity to the button within the time interval~$[a, b]$ so that the button is pressed and the pan remains hot. A CBF for $\phi$ is $B(x, t) = \epsilon^2 - \|x - x^*\|^2 + \gamma(t)$, where $\gamma(\cdot)$ controls the rate at which the robot reaches $x^*$ such that ${\gamma(b) = 0, B(x, a) \geq 0}$. For example, we can use ${\gamma(t) = \max\{0, \frac{\|x(a) - x^*\|^2}{a - b}(t - b)}\}$ so that $B(x, a) = \epsilon^2$. If there exists a control law $u(x, t)$ so that the solution of~\eqref{DS_form}  satisfies $B(x(t), t) \geq 0$ for all $t \in [a, b]$, then $\phi$ is satisfied. Since $\gamma(t)$ is monotonically decreasing until $t = b$, the size of the set $\mathcal{C}(t)$ defined in Section~\ref{sec:stl} will decrease until $t=b$ and the control law ${u(x, t) \in K(x, t)}$ will drive the state $x$ $\epsilon$-close to $x^*$. We can formulate a QP similar to~\cite{stl_lars} and compute $u(x, t)$ that renders the set $\mathcal{C}(t)$ forward invariant for the dynamical system~\eqref{DS_form} with the CBF $B(x, t)$: 
\begin{equation}
    \begin{aligned}
        \quad & u(x, t) = \argmin_{v} \quad \bigl\|v\bigr\|^2_2 \\
        \textnormal{s.t.} \quad & \frac{\partial B(x, t)}{\partial x}^{\top}\left(\hat{f}_{n(t)}(x) + v\right) + \frac{\partial B(x, t)}{\partial t}\geq -\gamma(B(x, t))
    \end{aligned}
    \label{CBF_opt}
\end{equation}
The above problem is always feasible if $B(x, t)$ is a valid CBF and~\cite{stl_lars} presents a closed-form expression under the additional assumption that $B(x, t)$ is either non-decreasing or non-increasing in $t$. Both problems~\eqref{safety_opt} and~\eqref{CBF_opt} are QPs that we solve efficiently in real-time using OSQP solver~\cite{osqp}. 

\subsubsection{Switching between behaviors}
\label{sec:switch}
The RTL specification requires the robot to switch between different behaviors depending on the changing environmental observations. Consider a scenario from our running example in Fig.~\ref{fig:aut_eg_small}, where the pan is hot ($h=1$) at all time $t \leq t_1$, but is not hot enough ($h=0$) at time $t_2 > t_1$. Then, the robot would be performing the \textit{stirring} motion at time $t \leq t_1$ which involves commanding the virtual control $u(x)$ computed from~\eqref{safety_opt} with $\hat{f}:=\hat{f}_{n(t)}$ modeling the stirring motion. Now, at time $t_2$, since $h=0$, the robot should follow the \textit{press button} behavior to satisfy $\Psi$, which could be satisfied by commanding $u(x, t)$ from~\eqref{CBF_opt}. This scenario involves switching between two types of motion planning strategies that correspond to $p_m = s$ and $p_m = p$. 

\textbf{Mixing strategy}: We propose a mixing strategy that connects two successive motion planners when the robot has to switch between behaviors so that there is a smooth transition. Let the robot follow a behavior $a \in \mathcal{A}_c$ during the time $t \leq t_1$ and that it has to switch to behavior $b \in \mathcal{A}_c$ when time $t > t_1$. Let the nominal vector field be $\hat{f}_a(x)$ for behavior $a$ and $\hat{f}_b(x)$ for behavior $b$. Let the virtual control input be $u_a(x, t)$ for behavior $a$ and $u_b(x, t)$ for behavior $b$. Then, the commanded reference velocity at time $t_1$ is $\dot{x}_{\textnormal{ref}}(t_1) = \hat{f}_a(x) + u_a(x, t_1)$. We define the mixing strategy when the robot switches from behavior $a \to b$ at time $t_1$ as
\begin{alignat*}{2}
&\dot{x}_{\textnormal{ref}}(t) &&= (1 - \beta(t))\dot{x}_{\textnormal{ref}}(t_1) + \beta(t)\left(\hat{f}_b(x) + u_b(x, t)\right), \\
& \beta(t) &&= \max\left(0, \min\left(1, \frac{t - t_1}{\Delta t}\right)\right),
    \label{mix}
\end{alignat*}
where $\Delta t$ is the during for which the mixing strategy is implemented. The mixing factor~$\beta(t)$ linearly increases from $0$ to $1$ over the time interval $t \in [t_1, t_1 + \Delta t]$ so that the commanded reference velocity smoothly varies from $\dot{x}_{\textnormal{ref}}(t_1)$~(behavior~$a$) to $\dot{x}_{\textnormal{ref}}(t_1 + \Delta t)$ (behavior $b$). If we increase $\Delta t$, the robot will take longer to switch between behaviors and vice-versa. 
\section{Experimental Results}
\label{sec:exp_results}
We demonstrate our method on a Franka robot arm performing two experiments: (i) wiping a whiteboard and (ii) wiping a mannequin, both while adapting to environmental changes and being compliant to human interactions.   
\subsection{Wiping a whiteboard}
\label{sec:exp_whiteboard}
The task execution by the robot using our method is given in Fig.~\ref{fig:board_snaps_exp}, for the same scenario that was introduced in Fig.~\ref{fig:board-snaps}. On a high level, the robot should switch between two different nominal wiping motions, react to the environment by wiping off any stain on the board, and adapt to external disturbances such as the eraser being dropped off from the end effector. The uncontrollable propositions are (i)~${eraser=1}$ $\Leftrightarrow$ the end-effector is holding an eraser, (ii)~${stain} = 1\Leftrightarrow$~there is a stain on the whiteboard, (iii)~$left = 1 \Leftrightarrow$ nominal wiping motion on the left side of the board and (iv) \textit{right} = 1 $\Leftrightarrow$ a different wiping motion on the right side of the board. The boolean values for \textit{left} and \textit{right} are controlled by the user using a GUI, the stain is detected by an Intel Realsense camera that returns the bounding box of the stain, and we assume the eraser is dropped if the gripper of the end-effector is open. The controllable propositions that model different desired behaviors are given in Table~\ref{tab:controllable_board}. The RTL specification that the robot should satisfy is
\begin{equation*}
\resizebox{0.98\linewidth}{!}{$\begin{aligned}
    &\Psi ::= && G\left(\left(eraser \ \textnormal{and} \ left \ \textnormal{and} \ \neg stain\right) \Rightarrow \ w_{\textnormal{left}}\right) \ \textnormal{and} \\
    & &&G\left(\left(eraser \ \textnormal{and} \ right \ \textnormal{and} \ \neg stain\right) \Rightarrow \ w_{\textnormal{right}}\right)) \ \textnormal{and} \\
    & &&\left(\left(eraser \ \textnormal{and} \ stain\right) \ \Rightarrow s_p\right) \  \textnormal{and} \
    \left(\neg eraser \Rightarrow e_p\right) \ \textnormal{and} \\
    & &&G\left(\left(eraser \ \textnormal{and} \ stain\right)\Rightarrow \left(s_p \ U \ w_{\textnormal{stain}}\right)\right).
\end{aligned}$}
\end{equation*}
% \begin{alignat*}{4}
%     &\Psi := && G\left(\left(eraser \ \textnormal{and} \ left \ \textnormal{and} \ \neg stain\right) \Rightarrow \ w_{\textnormal{left}}\right) \ \textnormal{and} \\
%     & &&G\left(\left(eraser \ \textnormal{and} \ right \ \textnormal{and} \ \neg stain\right) \Rightarrow \ w_{\textnormal{right}}\right)) \ \textnormal{and} \\
%     & &&\left(\left(eraser \ \textnormal{and} \ stain\right) \ \Rightarrow s_p\right) \  \textnormal{and} \
%     \left(\neg eraser \Rightarrow e_p\right) \ \textnormal{and} \\
%     & &&G\left(\left(eraser \ \textnormal{and} \ stain\right)\Rightarrow \left(s_p \ U \ w_{\textnormal{stain}}\right)\right).
% \end{alignat*}

\vspace{-10pt}
\subsection{Wiping a human mannequin}
\label{sec:exp_mannequin}
We present the experimental details of another wiping task, this time on a human mannequin, where the robot motions are given in Fig.~\ref{fig:mannequin_snaps}. The robot should switch between two different nominal motions on the legs and the hands, wipe the back if the human gestures in that region, and wet the towel at regular intervals where a human squeezes the towel after wiping. The uncontrollable propositions are: (i)~${leg =1~\Leftrightarrow}$~wipe the legs, (ii) ${hand = 1}~\Leftrightarrow$ wipe the hands, (iii)~${back = 1}~\Leftrightarrow$ human gesturing on the back, (iv)~${wet = 1\Leftrightarrow}$~towel is wet, and (v) {${human= 1\Leftrightarrow}$~human is inside the workspace to squeeze the towel. The boolean values for \textit{leg} and \textit{hand} are controlled by the user using a GUI, the human gesture on the back is detected by an Intel Realsense camera, and the human being inside the robot workspace is detected by Optitrack markers on the human hand. We assume that the towel becomes dry, i.e., ${wet = 0}$, every 30 secs. The controllable propositions are given in Table~\ref{tab:controllable_m}, and RTL specification is
\begin{equation*}
\resizebox{1\linewidth}{!}{$\begin{aligned}
    &\Psi ::= && G\left(\left(wet \ \textnormal{and} \ leg \ \textnormal{and} \ \neg back\right) \Rightarrow  w_{\textnormal{leg}}\right) \ \textnormal{and} \\
    & &&G\left(\left(wet \ \textnormal{and} \ hand \ \textnormal{and} \ \neg back\right) \Rightarrow w_{\textnormal{hand}}\right)) \ \textnormal{and} \\
    & &&G\left(\left(wet \ \textnormal{and} \ back\right) \Rightarrow w_{\textnormal{back}}\right) \  \textnormal{and} \
    G\left(\neg wet  \Rightarrow d_p\right) \ \textnormal{and} \\
    & &&G\left(\left(human \ \textnormal{and} \ wet\right)\Rightarrow b_p\right) \ \textnormal{and} \ G\left(\left(\neg human \ \textnormal{and} \ wet\right)\Rightarrow \texttt{true}\right).
\end{aligned}$}
\end{equation*}
% \begin{alignat*}{4}
%     &\Psi := && G\left(\left(wet \ \textnormal{and} \ leg \ \textnormal{and} \ \neg back\right) \Rightarrow  w_{\textnormal{leg}}\right) \ \textnormal{and} \\
%     & &&G\left(\left(wet \ \textnormal{and} \ \textnormal{and} \ \neg back\right) \Rightarrow w_{\textnormal{hand}}\right)) \ \textnormal{and} \\
%     & &&G\left(\left(wet \ \textnormal{and} \ back\right) \Rightarrow w_{\textnormal{back}}\right) \  \textnormal{and} \
%     G\left(\neg wet  \Rightarrow d_p\right) \ \textnormal{and} \\
%     & &&G\left(\left(human \ \textnormal{and} \ wet\right)\Rightarrow b_p\right) \ \textnormal{and} \\
%     & &&
%     G\left(\left(\neg human \ \textnormal{and} \ wet\right)\Rightarrow \texttt{true}\right).
% \end{alignat*}
\textbf{Implementation details}: For both the experiments, we use Spot~2.0~\cite{spot_2} to construct the automaton~$\mathcal{B}_{\Psi}$ that has 43~states,  269~transitions for the whiteboard task; and 64~states, 510~transitions for the mannequin task. The nominal DS~$\hat{f}_a$ in Table~\ref{tab:controllable_board} and~\ref{tab:controllable_m} are learned from 3 demonstrations for each motion using Neural ODEs~\cite{nawaz2023learning} that efficiently capture periodic motions. For wiping off the stain~($w_{\textnormal{stain}}$), we generate a nominal trajectory online modeled as~$\hat{f}_{\textnormal{stain}}$ that covers the bounding box of the detected stain, and the trajectory is tracked using CLFs. 
We compute the time within which the robot should satisfy STL tasks as $t^* = \frac{\|x(0) - x^*\|}{v_u}$, where ${(t^*, x^*) \in \{(t_s, x_s), (t_e, x_e), (t_d, x_d), (t_b, x_b)\}}$, $v_u$ is analogous to a uniform velocity for the robot and $x(0)$ is the state when the robot starts to follow the relevant behavior. We satisfy the STL tasks using time-varying CBFs as described in Section~\ref{sec:CBF_STL} with ${\gamma(t) = \max\left\{0, \frac{\left\|x(0) - x^*\right\|^2_2 \left(e^{-t} - e^{-t^*}\right)}{1 - e^{-t^*}}\right\}}$ and $\epsilon = 5 \ $cm. Such an exponential~$\gamma$ function generates a motion plan that is faster at the beginning with smoother convergence to the target point~$x^*$, when compared to a linear~$\gamma$ function which is typically used in prior work~\cite{event_STL, stl_lars}. We use $\Delta t = 0.67$ secs for the mixing strategy that generates smooth switching behaviors while also reacting quickly to environmental changes. Additional experiment videos available at~\url{https://sites.google.com/view/rtl-plan-control-hri}.
\vspace{-6pt}
\section{Conclusion \& Future Work}
\label{sec:limit_concl}
\vspace{-5pt}
We propose a modular reactive planner which integrates discrete temporal logic based plans and continuous DS-based motion plans that guarantees task satisfaction and stability while adapting to unforeseen disturbances and external observations. We demonstrate our method on the Franka robot arm for reactive wiping tasks involving safe human interactions.  

One limitation of our work is not incorporating deadlock scenarios~\cite{deadlock} when robots cannot make task progress. Our future work aims to address such issues by integrating Large Language Models~\cite{LLM} into LTL specifications that can act at a high level as a supervisory control~\cite{supervisory}. Future work will also investigate planning strategies to satisfy task specifications on $\mathcal{SO}(3)$ and $\mathcal{SE}(3)$ manifolds.  
\vspace{-5pt}
% \addtolength{\textheight}{-12cm}   % This command serves to balance the column lengths
% on the last page of the document manually. It shortens
% the textheight of the last page by a suitable amount.
% This command does not take effect until the next page
% so it should come on the page before the last. Make
% sure that you do not shorten the textheight too much.

%%%%%%%%%%%%%%%%%%%%%%%%%%%%%%%%%%%%%%%%%%%%%%%%%%%%%%%%%%%%%%%%%%%%%%%%%%%%%%%%

%%%%%%%%%%%%%%%%%%%%%%%%%%%%%%%%%%%%%%%%%%%%%%%%%%%%%%%%%%%%%%%%%%%%%%%%%%%%%%%%

%%%%%%%%%%%%%%%%%%%%%%%%%%%%%%%%%%%%%%%%%%%%%%%%%%%%%%%%%%%%%%%%%%%%%%%%%%%%%%%%
% \section*{APPENDIX}

% Appendixes should appear before the acknowledgment.

% \section*{ACKNOWLEDGMENT}

%%%%%%%%%%%%%%%%%%%%%%%%%%%%%%%%%%%%%%%%%%%%%%%%%%%%%%%%%%%%%%%%%%%%%%%%%%%%%%%%

\bibliographystyle{IEEEtran}
\bibliography{IEEEabrv, IEEEexample}

\end{document}